\documentclass{article} 
\usepackage{nips13submit_e,times}
\usepackage{hyperref}
\usepackage{url}

\usepackage{epsfig}
\usepackage{picinpar}
\usepackage{graphicx}
\usepackage{amsmath}
\usepackage{amsthm}
\usepackage{amssymb,bm}
\usepackage[ruled,vlined]{algorithm2e}
\usepackage[titletoc]{appendix}
\usepackage{float,caption,multirow}

\newcommand{\mean}{\operatornamewithlimits{mean}}

\newcommand{\bs}{\mathbf}

\title{LIME: A Method for Low-light IMage Enhancement}

\author{
Xiaojie Guo\\
State Key Laboratory Of Information Security\\
Institute of Information Engineering, Chinese Academy of Sciences\\
\texttt{xj.max.guo@gmail.com} \\
}

\nipsfinalcopy 

\begin{document}

\maketitle

\begin{abstract}
When one captures images in low-light conditions, the images often suffer from low visibility. This poor quality may significantly degrade the performance of many computer vision and multimedia algorithms that are primarily designed for high-quality inputs. In this paper, we propose a very simple and effective method, named as LIME, to enhance low-light images. More concretely, the illumination of each pixel is first estimated individually by finding the maximum value in R, G and B channels. Further, we refine the initial illumination map by imposing a structure prior on it, as the final illumination map. Having the well-constructed illumination map, the enhancement can be achieved accordingly. Experiments on a number of challenging real-world low-light images are present to reveal the efficacy of our LIME and show its superiority over several state-of-the-arts.
\end{abstract}

\section{Introduction}

\begin{figure}[t]
\begin{center}
\includegraphics[width=0.8\linewidth]{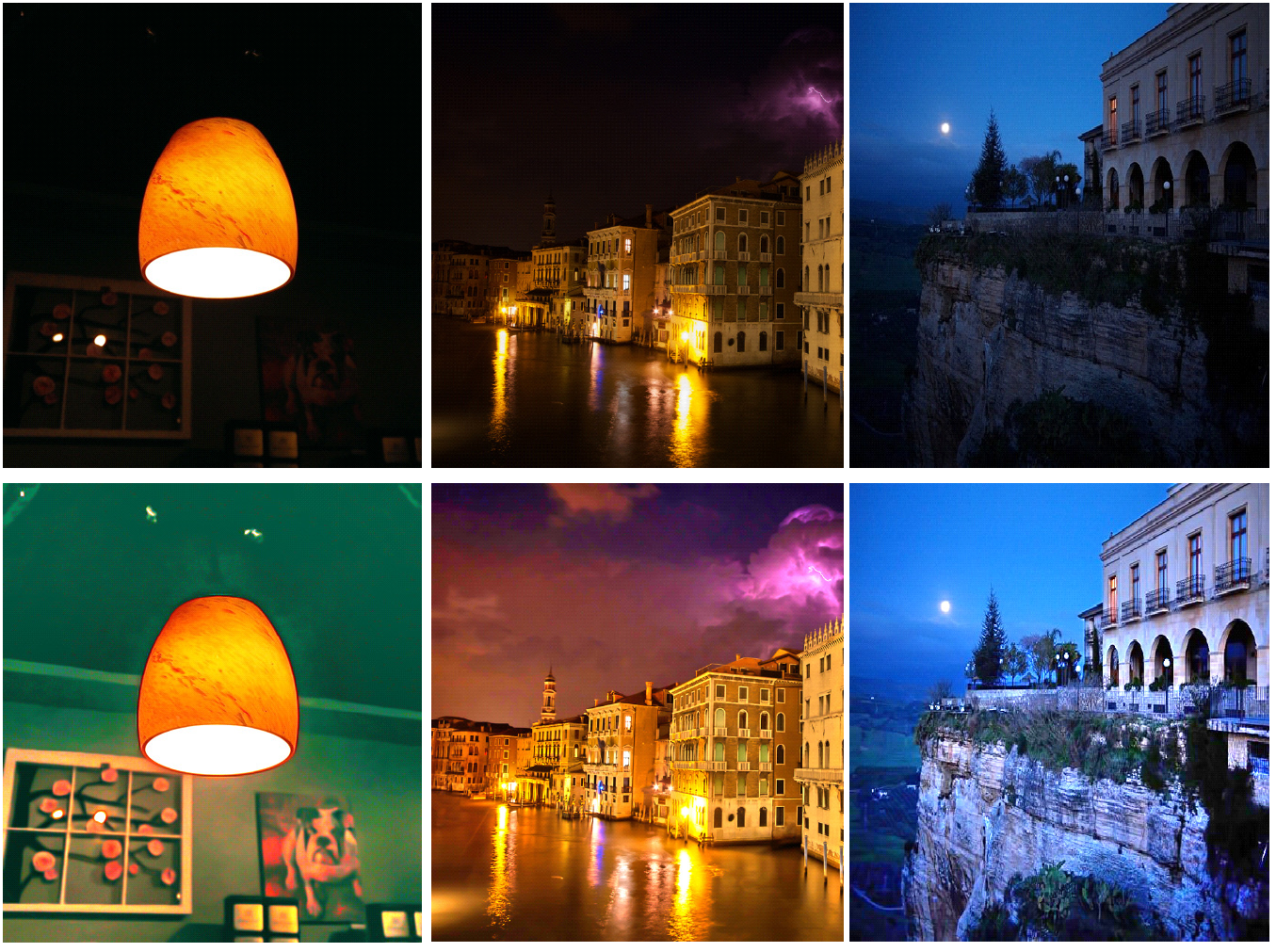}
\end{center}
\caption{\textbf{Upper Row:} Three natural low-light images. \textbf{Lower Row:} The enhanced results by our method.}
\label{fig:open}
\end{figure}

High-visibility images reflect clear details of target scenes, which are critical to many vision-based techniques, such as object detection \cite{ObjDet} and tracking \cite{Tracking}. But, images captured in low-light conditions are often of low visibility. Besides degrading the visual quality of images, it very likely hurts the performance of algorithms that are primarily designed for high-visibility inputs. Figure \ref{fig:open} provides three such examples, from which, we can see that many details, the paintings on the wall in the first case for example, have almost been ``buried" in the dark. To make the buried information visible again, low-light image enhancement is demanded.

Directly amplifying the low-light image is probably the most intuitive and simplest way to recall the visibility of dark regions. But this operation gives birth to another problem, say relatively bright regions might be saturated and thus loss corresponding details. Histogram equalization strategies \cite{AHE,FHE,DHE} can avoid the above problem by somehow forcing the output image to fall in the range $[0,1]$. However, in nature, they focus on contrast enhancement instead of exploiting real illumination causes, having the risk of over- and under-enhancement. The method proposed in \cite{NatEnh} tries to enhance contrast while preserving naturalness of illumination. Although it prevents the result from over-enhancement, in our test, its performance is not so attractive in both efficiency and visual quality.

\begin{figure}[t]
\begin{center}
\includegraphics[width=0.8\linewidth]{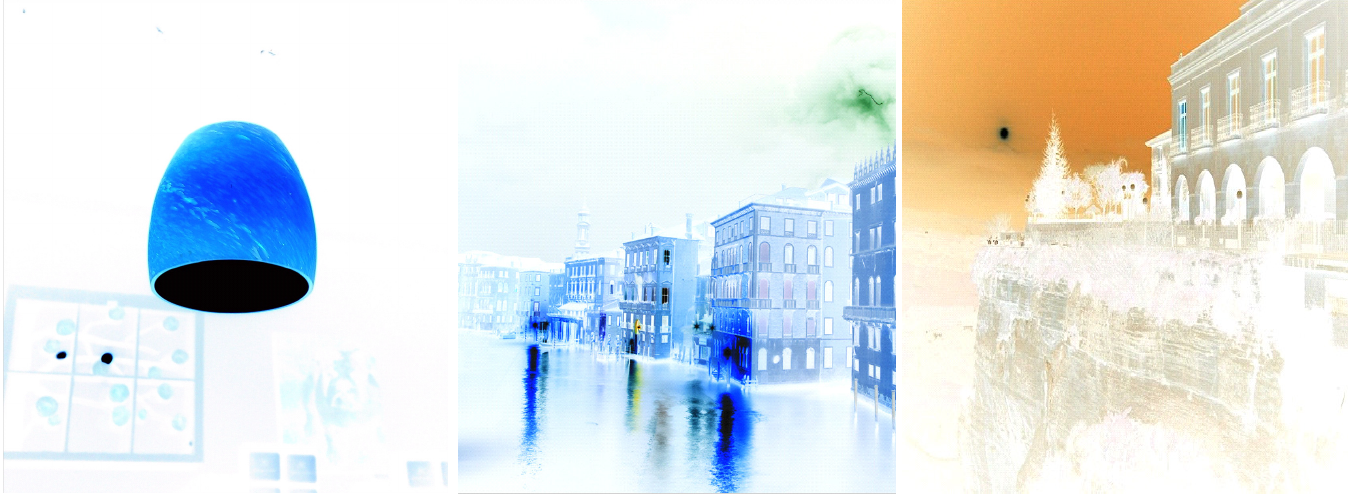}
\end{center}
\caption{The inverted versions (unrealistic images) of those shown in the upper row of Fig. \ref{fig:open}.}
\label{fig:ive}
\end{figure}

As noticed in \cite{LL11}, inverted low-light images look like hazy images, as shown in Fig. \ref{fig:ive}. Based on this observation, the authors of \cite{LL11} alternatively resorted to dehaze the inverted low-light images. After dehazing, the obtained unrealistic images is inverted again as the final enhanced results. Recently, Li \textit{et al.} followed this technical line and further improved the visual quality by first over-segmenting the input image and then adaptively denoising different segments \cite{LL15}. Even though the above methods can provide reasonable results, the basic model they rely on is lacking in physical explanation. This paper will try to connect this un-rooted model to a more physically meaningful one that our method adopts. 

This work intends to enhance a low-light image by estimating its illumination map. The illumination map is first constructed by finding the maximum intensity of each pixel in R, G and B channels. Then, we exploit the structure of the illumination and execute structure-aware smoothing to refine the illumination map. Experiments on a number of challenging images are conducted to demonstrate the advantages of our method in comparison with other state-of-the-art methods.

\section{Proposed Method}

Our method is built upon the following model, which explains the formation of a low-light image:
\begin{equation}
\bs{L} = \bs{I}\circ\bs{T},
\label{eq:dec}
\end{equation} 
where $\bs{L}$ and $\bs{I}$ are the captured image and the desired recovery, respectively. In addition, $\bs{T}$ represents the illumination map, and the operator $\circ$ means element-wise multiplication. The model \eqref{eq:dec} is with clear physical meaning, say the observed image can be decomposed into the product of the desired scene and the illumination image. As can be seen, the estimation of $\bs{T}$ is key to the recovery of $\bs{I}$. 

As mentioned, another widely used model is based on the observation that inverted low-light images $\bs{1}-\bs{L}$ look similar to haze images, which is thus expressed as \cite{HeDark,Effect,MengDehaze}:
\begin{equation}
\bs{1}-\bs{L} = (\bs{1}-\bs{I})\circ\tilde{\bs{T}}+a(\bs{1}-\tilde{\bs{T}}),
\label{eq:haze}
\end{equation}
where $a$ represents the global atmospheric light. Although the visual effect of inverted low-light images $\bs{1}-\bs{L}$ is intuitively similar to haze images, compared to the model \eqref{eq:dec}, the physical meaning of the above is not easy to directly explain. We will show the relation between \eqref{eq:haze} and \eqref{eq:dec} later.

\subsection{Initial Illumination Map Estimation}
\begin{figure}[t]
\begin{center}
\includegraphics[width=1\linewidth]{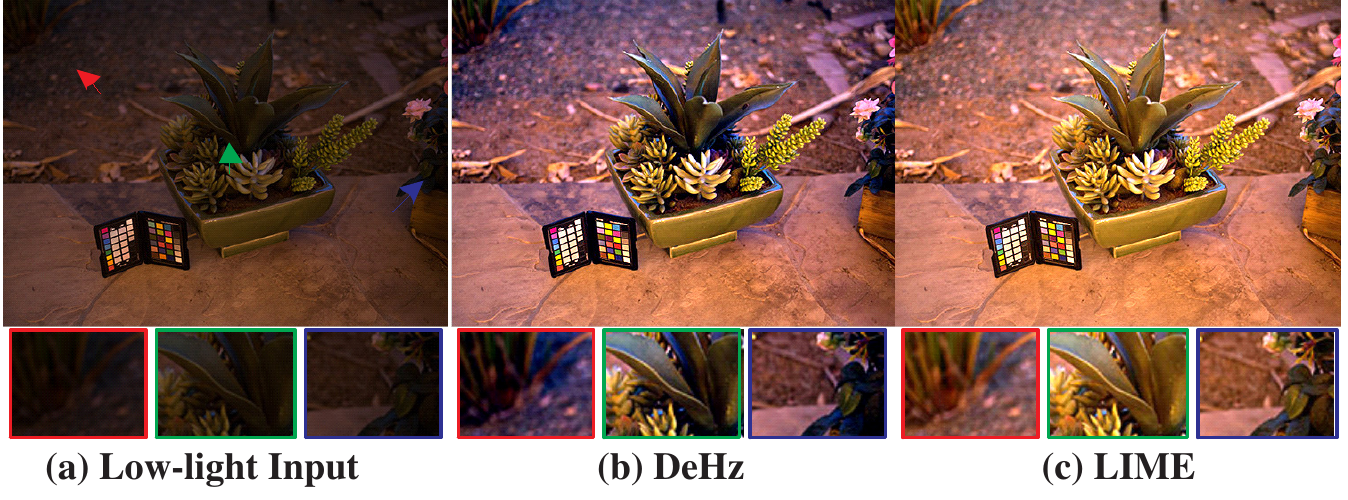}
\end{center}
\caption{Comparison of \eqref{eq:dehzI} and \eqref{eq:ourI} with the same illumination map. The atmospheric light $a$ estimated by \cite{HeDark} is larger than $0.95$. Even though, the difference is still noticeable.}
\label{fig:dif}
\end{figure}

As one of the first color constancy methods, Max-RGB \cite{Retinex} tries to estimate the illumination by seeking the maximum value of three color channels, say R, G and B. But this estimation can only boost the global illumination. In this paper, to handle non-uniform illuminations, we alternatively adopt the following initial estimation:
\begin{equation}
\hat{\bs{T}}(x) \leftarrow \max_{c\in\{R,G,B\}} \bs{L}^c(x).
\label{eq:ini}
\end{equation}
The obtained $\hat{\bs{T}(x)}$ guarantees that the recovery will not be saturated, because of
\begin{equation}
\bs{I}(x) = \bs{L}(x)/(\max_{c} \bs{L}^c(x)+\epsilon),
\label{eq:ourI}
\end{equation}
where $\epsilon$ is a very small constant to avoid the zero denominator. Let us here recall the dark channel prior, a commonly used prior to estimate the transmission map for dehazing \cite{HeDark}, on $\bs{1}-\bs{L}$ as follows:
\begin{equation}
\tilde{\bs{T}}(x) \leftarrow 1-\min_{c}\frac{1-\bs{L}^c(x)}{a}= 1-\frac{1}{a}+\max_{c}\frac{\bs{L}^c(x)}{a}.
\label{eq:dehzT}
\end{equation}
Accordingly, substituting \eqref{eq:dehzT} into \eqref{eq:haze} yields:
\begin{equation}
\bs{I}(x) = \frac{\bs{L}(x)-1+a}{(1-\frac{1}{a}+\max_{c}\frac{\bs{L}^c(x)}{a}+\epsilon)}+(1-a).
\label{eq:dehzI}
\end{equation}
We can see that when $a=1$, both \eqref{eq:ourI} and \eqref{eq:dehzI} reach the same result. But, if $a$ gets away from $1$, the equivalence between the model \eqref{eq:dehzI} \cite{LL11} and \eqref{eq:ourI} breaks (see Fig. \ref{fig:dif} for difference). 

In this work, we employ \eqref{eq:ini} to initially estimate illumination map $\hat{\bs{T}}$, due to its simplicity, although various approaches, like \cite{IR1,IR2,IR3}, have been developed to improve the accuracy in past decades. Most of these improvements essentially consider the local consistency of illumination by taking into account neighboring pixels within a small region around the target pixel. In the following, we provide a more powerful scheme to better achieve this goal.

\begin{figure*}[t]
\begin{center}
\includegraphics[width=1\linewidth]{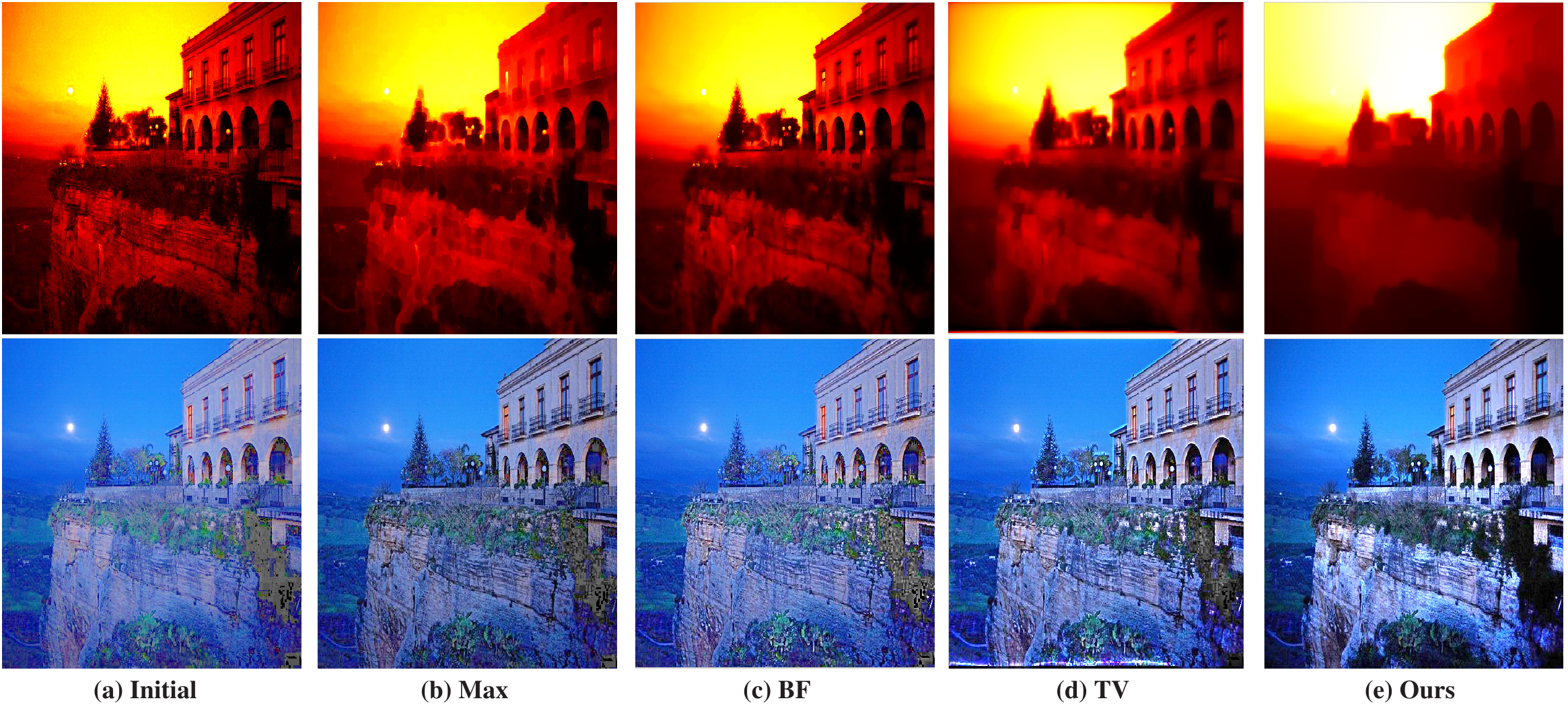}
\end{center}
\caption{Comparison of different illumination maps and corresponding enhanced results. \textbf{From (a) to (f):} Illumination map estimated individually on each pixel (Initial), refined by local max \eqref{eq:local} (Max), bilateral filtering (BF), $\ell_2$ loss total variation minimization (TV), and our structure-aware smoothing, respectively.}
\label{fig:map}
\end{figure*}

\subsection{Illumination Map Refinement}
As aforementioned, the illumination estimation can benefit from local consistency. Two representative ways are:
\begin{equation}
\begin{aligned}
\hat{\bs{T}}(x) \leftarrow& \max_{y\in\Omega(x)}~\max_{c\in\{R,G,B\}} \bs{L}^c(y);   \\
\hat{\bs{T}}(x) \leftarrow& \mean_{y\in\Omega(x)}~\max_{c\in\{R,G,B\}} \bs{L}^c(y),
 \end{aligned}
 \label{eq:local} 
\end{equation}
where $\Omega(x)$ is a region centered at pixel $x$, and $y$ is the location index within the region. These strategies can somewhat enhance the local consistency, but they are structure-blind. 

A ``good" solution should simultaneously preserve the overall structure and smooth the textural details. To address this issue, based on the initial illumination map $\hat{\bs{T}}$, we propose to solve the following optimization problem:
\begin{equation}
\min_{\bs{T}}\|\hat{\bs{T}}-\bs{T}\|_F^2 + \alpha\|\bs{W}\circ\nabla\bs{T}\|_1,
\label{eq:refM}
\end{equation}
where $\alpha$ ($0.15$ for all the experiments) is the coefficient to balance the involved two terms and, $\|\cdot\|_F$ and $\|\cdot\|_1$ designate the Frobenious and $\ell_1$ norms, respectively. Further, $\bs{W}$ is the weight matrix, and $\nabla$ that contains $\nabla_h{\bs{T}}$ (horizontal) and $\nabla_v{\bs{T}}$ (vertical), is the first order derivative filter. In the objective \eqref{eq:refM}, the first term takes care of the fidelity between the initial map $\hat{\bs{T}}$ and the refined one $\bs{T}$, while the second term considers the (structure-aware) smoothness. It can be seen that setting the weight matrix to $\bs{1}$ (all entries being $1$) leads \eqref{eq:refM} to a classic $\ell_2$ loss total variation minimization problem (TV for short) \cite{ADMTV}, which is also short of ability to distinguish between strong structural edges and texture \cite{RTV}.

Hence, the key is the design of $\bs{W}$. Inspired by RTV \cite{RTV}, for each location, the weight (\textit{e.g.} $\bs{W}_h(x)$) is set via:
\begin{equation}
\bs{W}_h(x) \leftarrow \sum_{y\in\Omega(x)}\frac{G_{\sigma}(x,y)}{|\sum_{y\in\Omega(x)}G_{\sigma}(x,y)\nabla_h\hat{\bs{T}}(y)|+\epsilon},
\end{equation}
where $G_\sigma(x,y)$ is produced by the Gaussian kernel with the standard deviation $\sigma$ (we use $2$ throughout this paper), and $|\cdot|$ is the absolute value operator. Please note that, different to RTV, our weight matrix is constructed based on the given $\hat{\bs{T}}$ instead of being iteratively updated according to ${\bs{T}}$. That means $\bs{W}$ only needs to be calculated once. 

Traditionally, the problem \eqref{eq:refM} can be effectively solved via alternating direction minimization techniques. To speed up the calculation, we approximate \eqref{eq:refM} by the following:
\begin{equation}
\min_{\bs{T}}\|\hat{\bs{T}}-\bs{T}\|_F^2 + \alpha\sum_x\frac{\bs{W}_h(x)(\nabla_h\bs{T}(x))^2}{|\nabla_h\hat{\bs{T}}(x)|+\epsilon}+\frac{\bs{W}_v(x)(\nabla_v\bs{T}(x))^2}{|\nabla_v\hat{\bs{T}}(x)|+\epsilon}.
\end{equation} 
As can be seen, the problem now only involves quadratic terms. Thus, the solution can be directly computed without requiring any iterations. Figure \ref{fig:map} shows a comparison of different methods on illumination map, from which, we can see the advance of our method.
\subsection{Other Operations}
\label{sec:post}

\begin{figure*}[t]
\begin{center}
\includegraphics[width=1\linewidth]{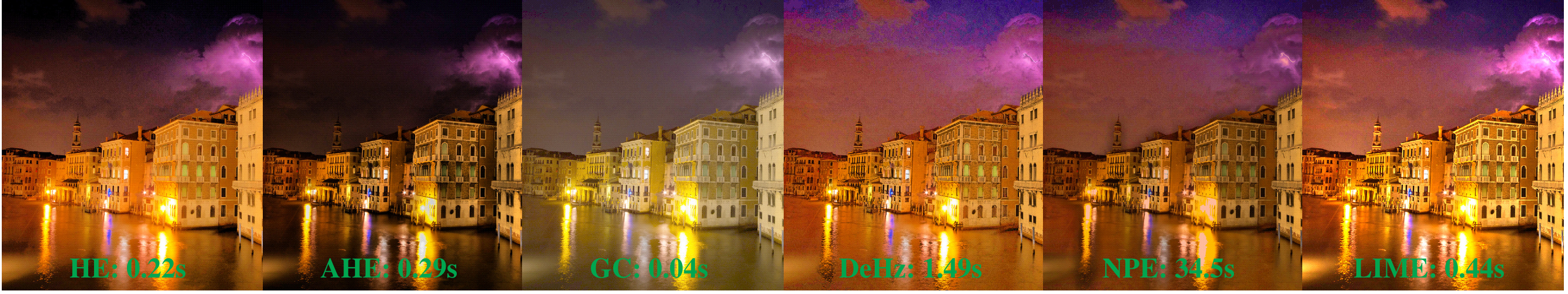}\\
\includegraphics[width=1\linewidth]{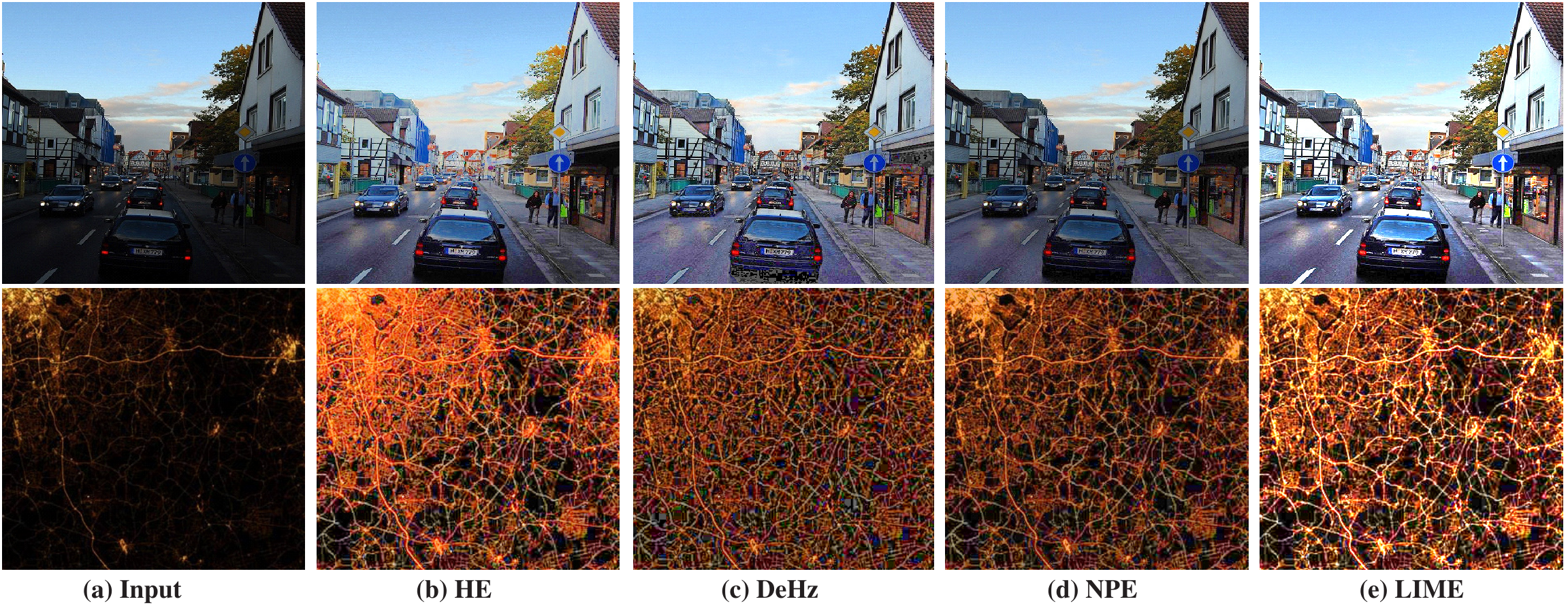}
\end{center}
\caption{Comparison of different methods without denoising involved.}
\label{fig:cmp1}
\end{figure*} 
\begin{figure}[!h]
\begin{center}
\includegraphics[width=0.8\linewidth]{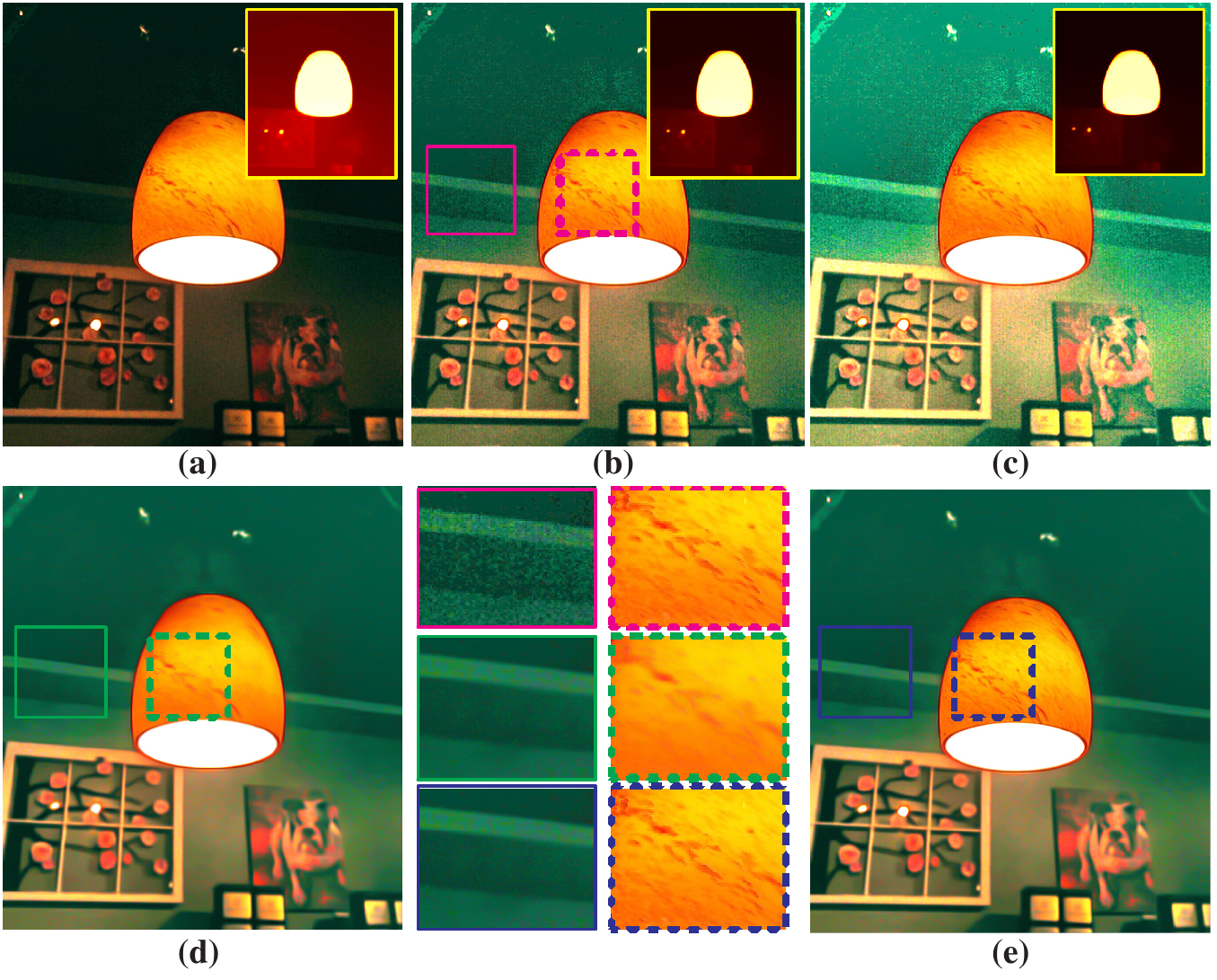}
\end{center}
\caption{Gamma correction, denosing and recomposition. (a)-(c) are the recovered images using $\bs{T}^{\gamma}$ with $\gamma=0.5$, $\gamma=0.8$ and $\gamma=1$, respectively. The corresponding illumination map is given in the up-right corner of each sub-picture. Noises appear in the enhanced images. (d) is the denoised version of (b) by BM3D, while (e) is the recomposed result of (b) and (e) by \eqref{eq:recp}. It can be seen from the zoomed-in patches that the recomposition adaptively keeps the fine details of the bright region and suppresses the noises of the dark region.}
\label{fig:finrec}
\end{figure}

Having the refined illumination map $\bs{T}$, we can recover $\bs{I}$ by following \eqref{eq:ourI}. One can also manipulate the illumination map through gamma transformation, say $\bs{T}\leftarrow\bs{T}^{\gamma}$. From the upper row of Fig. \ref{fig:finrec}, we can see the difference between the results by setting $\gamma$ to $0.5$, $0.8$ and $1$. For the rest experiments, we adopt $\gamma=0.8$. Moreover,  possible noises previously hiding in the dark are also accordingly amplified, especially for the very low-light inputs (regions), as shown in Fig. \ref{fig:finrec}. Denoising techniques are required to further improve the visual quality. Many off-the-shelf denosing tools, such as \cite{BM3D,NN,WNNM}, can be employed to do the job. Considering the comprehensive performance, BM3D \cite{BM3D} is the choice of this work. In our implementation, for further cutting the computational load, we only execute BM3D on the Y channel by converting $\bs{I}$ from the RGB colorspace into the YCbCr one. In addition, the magnitude of noises is not the same for different regions of the input, as the amplification is different. And BM3D treats different patches equally. Therefore, to avoid the unbalance of processing, \textit{e.g.} some (dark) places are well-denoised while some (brighter) over-smoothed, we employ the following operation:
\begin{equation}
\bs{I}_f \leftarrow \bs{I}\circ\bs{T}+\bs{I}_d\circ(\bs{1}-\bs{T}),
\label{eq:recp}
\end{equation}  
where $\bs{I}_d$ and $\bs{I}_f$ are the results after denoising and recomposing, respectively. The merit of this operation can be viewed from Fig. \ref{fig:finrec} (e), compared with Fig. \ref{fig:finrec} (d). We mention that the denoising, as a post-processing step, can be concatenated to any low-light image enhancing method. 
\begin{figure*}[t]
\begin{center}
\includegraphics[width=1\linewidth]{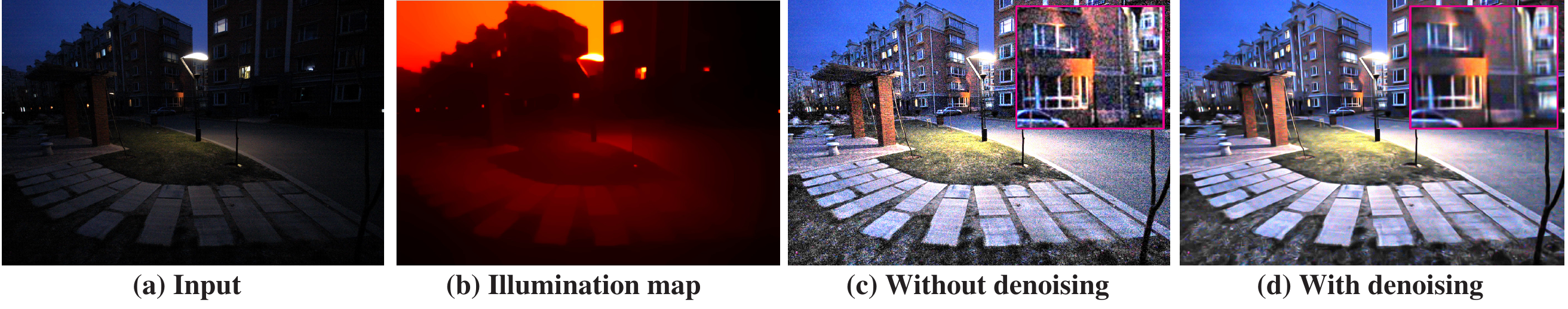}
\end{center}
\caption{Comparison of results without and with denoising.}
\label{fig:post}
\end{figure*}

\section{Experimental Results}
In this section, we compare our LIME with several state-of-the-art methods, including histogram equalization (HE), adaptive histogram equalization (AHE), Gamma Correction (GC), Dehazing based method \cite{LL11} (DeHz) and Naturalness Preserved Enhancement algorithm (NPE) \cite{NatEnh}. All the codes are in Matlab\footnote{HE and AHE uses \textit{histeq} and \textit{adapthisteq} functions integrated in the Matlab toolbox. GC is achieved by $\bs{L}^{\gamma}$, while the code of NPE is downloaded from the authors' website. The code of DeHz is not publicly available when this paper is prepared, but it is easy to be implemented based on \cite{HeDark}.}, which ensures the fairness of time comparison. All the experiments are conducted on a machine running Windows 7 OS with 64G RAM and 2.4GHz CPU.

Figure \ref{fig:cmp1} provides several comparisons. From the top row (the input is the second case of Fig. \ref{fig:open} with size $680$x$720$), we can see that AHE can not effectively recall the information in dark regions while GC ($\gamma=0.4$) changes the color of the whole image. These problems almost exist always, therefore we discard them for the rest comparisons. HE, DeHz and NPE outperform AHE and GC in this case, but are inferior to our method in terms of visual quality. In time cost, although LIME spends more than HE, AHE and GC, it is comparable to or even more efficient than DeHz, while much more efficient than NPE. Most cost of DeHz comes from the estimation of atmospheric light. Two more comparisons are given in Fig. \ref{fig:cmp1}, which are the additional evidence of the advantage of LIME, compared with HE, DeHz and NPE.

Figure \ref{fig:post} gives another test. The very low-light input hides intensive noises in the dark. After performing LIME, the details of the scene get enhanced, but the noises also come out, as shown in the middle of Fig. \ref{fig:post}. This is an inevitable problem encountered by almost all of existing low-light enhancement algorithms. As we have discussed in Sec. \ref{sec:post}, denoising is required. The right picture in Fig. \ref{fig:post} is the denoised result by executing BM3D on the middle of Fig. \ref{fig:post}, from which we can see the improvement in terms of visual quality. To allow more experimental verification and comparisons, we provide our code at \url{http://cs.tju.edu.cn/orgs/vision/~xguo/homepage.htm}

\section{Conclusion}
This paper has proposed an efficient and effective method to enhance low-light images for boosting the visual quality and offering contemporary vision applications with reliable inputs. The key to the enhancement is how well the illumination map is estimated. The structure-aware smoothing has been developed to improve the illumination consistency. The experimental results have revealed the advance of our method compared with several state-of-the-art alternatives.


\small{
\bibliographystyle{ieeetr}
\bibliography{sigproc.bib}
}

\end{document}